\documentclass[final,3p,times,twocolumn]{elsarticle}
\usepackage{booktabs}
\newcommand{\tabitem}{~~\llap{\textbullet}~~}

\usepackage{multirow}
\usepackage{blindtext, graphicx, amsmath, algorithm, algpseudocode, pifont, algcompatible, comment, layout, amsthm, amssymb}
\usepackage{enumitem}   
\usepackage{eso-pic}
\usepackage{booktabs}
\usepackage{float}

\usepackage[utf8]{inputenc}
\usepackage[english]{babel}
\usepackage{hyperref} 
\hypersetup{ colorlinks=true, linkcolor=black, filecolor=black, urlcolor=cyan, }

\usepackage{caption}
\usepackage{makecell}

\journal{ICT Express}

\begin{document}

\begin{frontmatter}

\title{AI and the Creative Realm: A Short Review of Current and Future Applications}
\author{Fabio Crimaldi\corref{cor1}}
\ead{crimaldi.1957381@studenti.uniroma1.it}

\author{Manuele Leonelli}
\ead{manuele.leonelli@ie.edu}

\address{Department of Psychology, University La Sapienza, Rome, Italy\\School of Science and Technology, IE University, Madrid, Spain}

\cortext[cor1]{Corresponding author}

\begin{abstract}
This study explores the concept of creativity and artificial intelligence (AI) and their recent integration. While AI has traditionally been perceived as incapable of generating new ideas or creating art, the development of more sophisticated AI models and the proliferation of human-computer interaction tools have opened up new possibilities for AI in artistic creation. This study investigates the various applications of AI in a creative context, differentiating between the type of art, language, and algorithms used. It also considers the philosophical implications of AI and creativity, questioning whether consciousness can be researched in machines and AI's potential interests and decision-making capabilities. Overall, we aim to stimulate a reflection on AI's use and ethical implications in creative contexts.
\end{abstract}


\end{frontmatter}

\section{Introduction}\label{sec1}

Creativity and AI are not two concepts that have been linked for a long time. Be it because of actual possibilities of the past technology, culture, or just intuition, AIs have never been seen or conceived by public opinion as a tool to make art or to generate new ideas. Public opinion has perceived AIs as tools that cannot create anything novel \citep{vincent2021integrating}. Nonetheless, the next concept of AI should achieve and surpass the possibilities of human intelligence in the creative field. AIs creating art have long been unthinkable in actionable terms because of the need for the right technology. Now, this is possible not only from a theoretical stance but also from a practical one.

The development of \textit{strong AIs} and the massification of the tools through which human-computer interaction can generate art has shifted the general public's perception of the matter \citep{kurt2018artistic}. Various studies have highlighted how AIs can eventually be used in creative contexts opening new pathways of research and application \citep{ameen2022toward,anantrasirichai2022artificial,mazzone2019art}.  Generative AI models are now extensively used for the most diverse goals (see Section \ref{sec:types} for an overview), and their use and capabilities will grow exponentially.  \textit{ChatGPT}, \textit{DALLE-2}, and \textit{Stable diffusion} are just a few examples of the broad environment now thriving with the novel computational capacities and possibilities we are and will be experiencing \citep{gozalo2023chatgpt}. 

The question of creativity and AI also has a philosophical side, whether creativity is uniquely linked with the mere fact of being a human being or is associated with the very deed of creating something that has never been seen or generated before \citep{sawyer2011explaining}. In \citep{boden_1998}, within the pioneering Sternberg's "Handbook of Creativity," \citep{sternberg1999handbook} there is an attempt to assess the existing kinds of creativity, describing them and evaluating how well AIs can perform in various creative contexts. The Handbook provides the foundations for the conceptualization of creativity and its study from an analytic viewpoint, away from the traditional cliches around it \citep{alves2022creativity,gerwig2021relationship,tang2019fostering,wu2020systematic}.
 
 The main question remains: Can we talk about more than the mere aggregation of data concerning AIs creations? Investigating this question is not this study's primary goal since we have yet to answer such a complex and multifaceted question. What is interesting is the possibility this question gives us to investigate conscience and whether this concept, historically always applied to humans and animals at most, could also be researched in machines and to what extent.

 It is essential to highlight how quantum computing could be a game-changer in the environment of AI due to the speeding up of the computational processes on a potentially large scale \citep{kurt2018artistic}. Such a technology could ensure a computational capability to generate and use a human-like AI soon  \citep{rawat2022quantum}. Estimates confirm that quantum computers will perform millions of times faster than the ones we currently use; the development of quantum computers and AIs are intertwined and could be the only way a strong AI could be developed, outpacing any brilliantly functioning human brain \citep{kurzweil2014singularity}.
 
One main detail that has to be taken into account is undoubtedly the use and potential set of interests that an AI could be foreseen to cover. This will constitute a significant factor in the experimentation with these technologies, setting new boundaries in what is acceptable and what is not acceptable that a computer could be doing or deciding \citep{vuarzaru2022assessing}. Here we provide novel categorizations of current uses and technologies for AI in the arts. Such categorizations will foster the identification of future trends and gaps to fill in future research on AI applications of creativity. In a nutshell, this study aims to stimulate a reflection on the functionality and goodness of AI usage in creative contexts. To date, it is the first review to consider the \textit{kind of art} created by AIs, the \textit{programming languages}, and \textit{algorithms} used for this goal.

\section{AI and Artistic Sensibility}
It is possible to classify AI into two categories, each with its peculiarities and use: \textit{Weak AI} and \textit{Strong AI} \citep{strong2016applications}.

\subsection{Weak AI}

Weak AI, also called “\textit{Narrow AI},” uses big data to gather and make sense of a vast amount of information in a specific application domain; this kind of AI often has proven not to pass the Turing Test \citep{boyd2011six}. Weak AIs can outperform or equal humans in a specific activity but cannot face new challenges in novel settings without a prior re-training  \citep{kaplan2019siri}. The use of these AIs in recent times is astonishing; virtually every single AI we can use right now in our daily life and that is open to public access is a weak AI \citep{ameen2022toward}. 

\subsection{Strong AI}
A strong AI, or "\textit{Artificial General Intelligence}," even though this latest term has been used differently in previous research \citep{fjelland2020general}, is an AI that has been trained to reason flexibly and holistically, basically like a human 
\citep{huang2018artificial}.  It conserves, though, access and computational capability unprecedented in human history. The mere concept of a strong AI entails connections and links between the knowledge that could be acquired, without any prior rigid specification of the context of use, taking into account the data of the environment dynamically and functionally, without specific prior training \citep{fjelland2020general}. 

The development of strong AIs, sometimes referred to as purely theoretical, is still in progress and is finding plenty of technical difficulties \citep{huang2018artificial}. Nonetheless, it has been noted how, for a couple of years, AIs have become able to perform creative tasks, sometimes with a high degree of credibility \citep{huang2018artificial}.  AIs can achieve this by considering contextual data and blending them with some taste to create something entirely new, possibly without input from a human being. Indeed, we are still not considering strong AIs for these creative tasks, but it is fascinating to think how the standard could be set higher whether such a development occurs. The critical asset of a strong AI is that it can learn by itself and has no limitations of tiredness and biases in its reasoning; the reasoning per se has to be holistic and full of correlations between the elements taken into account \citep{strong2016applications}.

This means that strong AIs can find patterns of connection and use of knowledge holistically, which stands for passing knowledge from one field to another after careful consideration based on the prospective efficiency of such a practice \citep{strong2016applications}. It is helpful to highlight that Strong AI is still just a theoretical concept and that no Strong AI, for the time being, has ever been developed, nor are we close to it \citep{fjelland2020general}. Indeed, the mere concept of developing a strong AI has raised much criticism. Many researchers still think such an achievement is impossible, at least in the way it was initially proposed \citep{fjelland2020general,fortnow2021fifty}.

\subsection{AI Consciousness and Artistic Sensibility}

There are two ways AI combines elements and harmonizes them to create what is stated to be “art”: symbolism and connectivism \citep{tao2022new}.  \textit{Symbolism} is a top-down method in which machines simulate the processes generally used in human cognition to depict something through rules of interpretation \citep{cardon2018neurons}. \textit{Connectivism} is a bottom-up process that states how creation can be done through the connection of various elements \citep{tao2022new}, simulating an artificial neural network through deep learning processes 
 \citep{aguera2017art}. This last approach needs enormous amounts of data to create an answer for the prompt presented and make sense of the given stimuli  \citep{zylinska2020ai}.

 Unfortunately, the fact that all the AIs we are currently using are so-called Weak AIs implies that there is no sensibility or holistic application of knowledge in what AIs do. The term "Artistic sensibility" is defined in this work as “\textit{the sensitivity and capacity to appreciate and act upon concerns of or pertaining to art and its production}” \citep{ingman2022artistic} and not the mere capacity to distinguish different art styles and peculiar influences in a creation. However, it could be argued that the concept of artistic sensibility, primarily when related to a machine, is unclear in meaning. The variables that come into play and make up the artistic creation are still not isolated. Furthermore, it is still undetected whether a massive shift in creative practices using AIs as tools could lead to a general homologation of the works of art \citep{manovich2018ai}. More studies are needed to properly understand what kind of artistic sensibility could result from machines applied to art and which developments arise from them. 
 
\section{AI in the Arts}

\subsection{Fields of Application}
\label{sec:types}
AIs have been proven to have the possibility of being applied in different fields of expression, conveying different kinds of messages in different styles. Extensive research has been led about how AIs can be applied in creative deeds, even in collaboration with highly creative humans, and their consequences. Table \ref{table:art} summarizes significant achievements and future challenges in the main artistic areas of application of AI.

\begin{table*}[]
    \centering
    \scalebox{0.75}{
    \begin{tabular}{lll}
    \toprule
   \textbf{Kind of art}      & \textbf{Major achievements} & \textbf{Future implications} \\
   \midrule
  \multirow{5}{*}{\makecell[l]{\textit{Creating images} \\\citep{cetinic2022understanding,jadav2022evolution,lyu2022communication,mazzone2019art,miller2019artist}}  }     &  \tabitem  \makecell[l]{Viewers genuinely liking the artworks and\\  engaging in conversations about the process} & \tabitem \makecell[l]{More sophisticated conceptual work\\ as more artists explore AI tools} \\
   &  \tabitem \makecell[l]{Artistic style classification, object detection,\\ and similarity retrieval}  & \tabitem \makecell[l]{Fruitful partnership between artists and\\ creative AI systems}\\
  & \tabitem \makecell[l]{Development of a machine process that\\ focuses on understanding the process of creativity}& \tabitem \makecell[l]{Democratization of art: more
opportunities\\ for artists with less resources}\\
  & \tabitem Generation of images from text & \tabitem \makecell[l]{Preservation and restoration of art}\\ 
  & \tabitem Viewers not recognizing
the artworks as AI-derived\\
  \midrule
  \multirow{4}{*}{\makecell[l]{\textit{Creative text writing} \\ \cite{bajohr2022paradox,gunser2021can,hitsuwari2023does,kobis2021artificial,mirowski2022co,schober2022passing,woo2022student,yang2022ai}} } & \tabitem \makecell[l]{Incapability of reliably detecting\\
algorithm-generated poetry} & \tabitem \makecell[l]{New domains, such as crafting artificial \\ online reviews, patent claims, or fake tweets} \\
& \tabitem \makecell[l]{Understanding the potential implications\\ of language-generation algorithms}
 & \tabitem \makecell[l]{Experimental studies examining how people\\ react to algorithm-generated text}\\
& \tabitem \makecell[l]{Role of human input in algorithm implementation}& \tabitem Impact on questions related to authorship\\
&& \tabitem Emergence of
CryptoArt \\
\midrule 
\multirow{3}{*}{\makecell[l]{\textit{Music} \\\cite{chu2022empirical,civit2022systematic,hernandez2022music,hong2022human,moysis2023music}} }  & \tabitem \makecell[l]{Building interactive models that allow\\ composers to interact with AI}  & \tabitem \makecell[l]{Developing a general subjective evaluation\\ method for measuring creativity}\\
 &\tabitem \makecell[l]{Adapting music
composition to the majority\\ of the current musical structure} & \tabitem \makecell[l]{Defining what makes a composition\\ different from others}\\
 & \tabitem Mimicking styles of various composers successfully
& \tabitem \makecell[l]{Better modeling of long-term relationships\\ in time and harmony axes}\\
\midrule
\multirow{3}{*}{\makecell[l]{\textit{Sculpture} \\\cite{bidgoli2019deepcloud,dube2021ai,ge2019developing,koutsomichalis2018generative,lehman2016creative}} }  & \tabitem \makecell[l]{Autoencoders and GUI for manipulation of\\ high-dimensional latent spaces}  & \tabitem \makecell[l]{Introduction of physical constraints\\ in the AI creative process}\\
 & \tabitem New geometric models for sculpture and design
& \tabitem \makecell[l]{Integration of  semantic
segmentation}\\
\midrule 
\multirow{3}{*}{\makecell[l]{\textit{Videos} \\\cite{fei2021exposing,jafar2020forensics,kim2020tivgan,li2018video,masood2022deepfakes,singer2022make}} }  & \tabitem \makecell[l]{High accuracy in video manipulation detection}  & \tabitem \makecell[l]{New techniques for identification and\\ prevention of digital misinformation}\\
 &\tabitem \makecell[l]{Detection of face morphing attacks and \\ 3D lighting inconsistencies} & \tabitem \makecell[l]{Advances in the field of entertainment\\ using computer vision and image processing}\\
 & \tabitem Learning of detailed face reconstruction
& \tabitem \makecell[l]{Use of multi-task pre-training \\ for improved visual synthesis tasks}\\
 \bottomrule
    \end{tabular}}
    \caption{Major achievements and future implications in the main artistic areas of application of AI.}
    \label{table:art}
\end{table*}

Furthermore, AIs have also proven to be particularly effective in not-so-specific fields of application, such as communication or marketing, that entail a lot of different skills and strategies to convey a message in a very different way to what is done in an artistic setting \citep{ameen2022toward}. This semi-creative use of such technologies is regarded as the future of communication practices and content creation \citep{schiessl2022artificial}, which are still regarded as creative while not being considered purely artistic.

\subsection{Programming Languages}

\begin{table*}[]
    \centering
    \scalebox{0.8}{
    \begin{tabular}{lll}
    \toprule
   \textbf{Programming language}      & \textbf{Advantages} & \textbf{Disadvantages} \\
   \midrule
  \multirow{3}{*}{\makecell[l]{\textit{Java}\\ \citep{neumann2002programming,sharma2015improved,sharma2021artificial}}}       &  \tabitem  Large community and resources & \tabitem Slow performance \\
   &  \tabitem Object-orientation & \tabitem High memory usage\\
  & \tabitem Platform independence & \tabitem Verbosity\\
  \midrule
  \multirow{4}{*}{\makecell[l]{\textit{Python}\\\citep{charniak2014artificial,rozov2020,sharma2015improved}}} & \tabitem Easy to learn & \tabitem Slow performance \\
& \tabitem Large community and resources & \tabitem Poor memory management\\
& \tabitem Versatility & \tabitem Dynamic typing \\
& \tabitem Easy integration with other languages & \tabitem Limited parallel processing \\
\midrule 
\multirow{3}{*}{\makecell[l]{\textit{C++}\\ \citep{belson2020,naveed2021,neumann2002programming}}} & \tabitem High performance & \tabitem Complex syntax \\
 & \tabitem Control over memory management & \tabitem Lack of built-in features\\
 && \tabitem Lack of flexibility\\
 \midrule
 \multirow{2}{*}{\makecell[l]{\textit{LISP}\\ \citep{adetiba2021,hashimoto2021,sharma2021artificial}}} & \tabitem Expressiveness
& \tabitem Limited libraries\\
 & \tabitem Symbolic processing& \tabitem Slow Performance \\
 \bottomrule
    \end{tabular}}
    \caption{Advantages and disadvantages of the most common programming languages for AI in the arts.}
    \label{tab:languages}
\end{table*}

Regarding the programming of creative AIs, Python is the most commonly used language for programming creative AIs, generally for its visible and functional versatility in use and application  \citep{charniak2014artificial}. At the same time, other popular programming languages for creative AI goals have been used, such as Java, C++, and LISP \citep{sharma2021artificial}. Table \ref{tab:languages} summarizes the main advantages and disadvantages of each of these languages (based on \citep{neumann2002programming,sharma2015improved}). 

\subsection{Algorithms}

\begin{table*}[]
    \centering
    \scalebox{0.8}{
    \begin{tabular}{lll}
    \toprule
   \textbf{Algorithm}      & \textbf{Advantages} & \textbf{Disadvantages} \\
   \midrule
  \multirow{4}{*}{\makecell[l]{\textit{Genetic algorithms} \\\citep{adamik2022fast,dalle2022job,spence2022gastrophysics,yang2020nature}}  }     &  \tabitem  Ability to handle complex multi-faceted problems & \tabitem Slow convergence \\
   &  \tabitem Novel and unexpected solutions & \tabitem Risk of premature convergence\\
  & \tabitem Parallel processing & \tabitem Hard interpretation of solutions\\
  & \tabitem Handling of noisy and incomplete data\\
  \midrule
  \multirow{3}{*}{\makecell[l]{\textit{Neural networks} \\\citep{ge2019developing,karras2021alias,mazzone2019art,wang2022rewriting}} } & \tabitem Handling of  complex, nonlinear data & \tabitem Need of plenty of data and computational resources to train \\
& \tabitem Learning from data and generalizations
 & \tabitem Difficulty of interpretation\\
& \tabitem Handling of noisy data & \tabitem Hard to train and optimize \\
\midrule 
\multirow{4}{*}{\makecell[l]{\textit{Decision trees} \\\citep{alonso2020teaching,basysyar2022prediction,lin2013data,meza2019predictive}} }  & \tabitem Ease of interpretation & \tabitem Limited accuracy \\
 & \tabitem Easy to train and learn & \tabitem Prone to Overfitting\\
 &\tabitem Versatility for many applications & \tabitem LUnstable solutions\\
 & \tabitem No data pre-processing
& \tabitem Risk of bias\\
 \bottomrule
    \end{tabular}}
    \caption{Advantages and disadvantages of the most common algorithms for AI in the arts.}
    \label{tab:algo}
\end{table*}

An extensive range of algorithms has been extensively tested and employed in the creative
application of AI. These are summarized in Table \ref{tab:algo}. In this context, it is crucial to zoom in on the most widely utilized
algorithms, considering their strengths and weaknesses. Engaging in discussions
about these AI algorithms in the creative field holds significant value as it catalyzes
innovation, amplifies creative potential, and cultivates collaborative endeavors.
Simultaneously, such discussions offer valuable insights into the future of artistic expression
by exploring the intersection of technology and art. By bridging the gap between disciplines,
these conversations empower artists to embark on co-creation journeys with AI systems,
facilitating the exploration of uncharted territories and pushing the boundaries of
conventional art forms.

\section{Discussion}
In light of the legitimate concerns surrounding the future of the arts, it is essential to pay
attention to certain key elements to delve deeper into the subject and gain a
comprehensive understanding of it. First of all, no AI will ever conceive something new, not that we know of and not that we could ever imagine  \citep{vincent2021integrating}. For now, creating something entirely unseen from an artistic perspective through AI is impossible  \citep{dwivedi2021artificial}. In this regard, the creative industry will still need human professionals to work well. Secondly, AI tools cannot be used to change perspectives on how societal stereotyping works. They cannot try to change it or conceive something outside those lines, yielding an inevitable frustration and possibly causing a backlash in many fields, i.e., communication agencies, especially when discussing rapidly changing societal topics  \citep{fessler}. What has been noticed is that generally, the jobs that are taken from an AI more easily are low-level and manual, basically, those that do not require a lot of studying and professional training \citep{wang2018living}. For now, the creative realm remains a human prerogative.

\subsection*{Current Challenges}

Despite the potential benefits of using AI in the arts, several challenges are common across all fields of application. One challenge is balancing creativity and control \citep{galanter2019artificial}. AI-generated art can sometimes lack the intentionality and emotional resonance of human-made art, and finding the right balance between the artist's input and the AI's output can be tricky \citep{galanter2019artificial}. 
Another challenge is the need to train and fine-tune the generative models to produce high-quality outputs and not use biased data, which could eventually result, for instance, in homophobic or racist results \citep{mazzone2019art}. Another final challenge implies the concept of creativity per se, what is creative and what is not, and whether that could be linked with original artistic value \citep{boden_1998}.

\subsection*{What is next in creative AI?}

AI creativity is a domain that is experiencing swift progress, and the forthcoming future holds both thrilling opportunities and uncertainties. The generative AI revolution is still in its initial phases. The future is expected to be more unpredictable than in the recent past, possibly bearing more revolutions in the forthcoming years, especially from a methodological point of view \citep{dahlin2021mind}. This instability is prompted by AI developers competing to design and implement progressively more powerful digital intellects that nobody can foresee or adequately manage \citep{cetinic2022understanding}. As digital content and delivery channels continue infiltrating all artistic forms and expressions, AI's role in creativity is undeniably expanding \citep{dahlin2021mind}.

Within the following decade, AI is projected to play a critical function in the creative process, based on the view of almost 50\% of professionals in the creative sector \citep{tanjil}. This involves exploring the potential for AI to facilitate upcoming-generation consumer experiences such as the metaverse and cryptocurrencies \citep{sabry2020cryptocurrencies}. Nevertheless, as AI technology advances, there are concerns that it could jeopardize human autonomy, agency, and skills \citep{prunkl2022human}. A significant risk would be given by the limitations a vast reliance on these technologies could yield \citep{laitinen2021ai}.

Today, AI can only partially replace what a top-level human creator could be conceiving, as already noticed for advertising communication campaigns \citep{dwivedi2021artificial}. It is worth noting that we are just at the start of this technology and that computing power and algorithms will be and are already being developed to better the situation \citep{dwivedi2021artificial}. In this context, the main ontological difference between AI and human intelligence is that while the first is driven by meaningful purpose, the second creates through divergence of what has been established earlier, thus not giving a new stance on the matter of creation \citep{lehnert2014advertising}.

In conclusion, the future of AI creativity is anticipated to be characterized by swift growth and expansion. Within the next ten years, AI will likely assume a significant role in the creative process, transforming several artistic professions and giving rise to novel consumer experiences.
Nonetheless, as AI evolves, there are anxieties regarding its potential threats to human autonomy and skills \citep{calvo2020supporting}. The generative art and AI field face many challenges, ranging from technical difficulties in physically creating the artwork to overcoming bias in training data and capturing the nuances of human expression and emotion in creative works. However, with continued research and development, AI has the potential to revolutionize the creative process and expand the possibilities of art.

\section*{Conflict of interest}
The authors declare that there is no conflict of interest in this paper.



 \bibliographystyle{elsarticle-harv}
\bibliography{sample.bib}
\end{document}